\documentclass[letterpaper, 10 pt, conference]{ieeeconf}  

\IEEEoverridecommandlockouts                              

\usepackage{graphics} 
\usepackage{epsfig} 
\usepackage{times} 
\usepackage{amsmath} 
\usepackage{amssymb}  
\usepackage[normalem]{ulem}

\let\labelindent\relax
\usepackage{enumitem}

\usepackage{xcolor}

\usepackage{multirow}
\usepackage[normalem]{ulem}
\usepackage{cite}

\title{\LARGE \bf
Data-Driven Safety Verification for Legged Robots
}

\author{Junhyeok Ahn, Seung Hyeon Bang, Carlos Gonzalez, Yuanchen Yuan, and Luis Sentis
\thanks{J. Ahn is with the Department of Mechanical Engineering, The University of Texas at Austin, TX, 78712, USA. Email: junhyeokahn91@utexas.edu}%
\thanks{S. H. Bang, C. Gonzalez, Y. Yuan, and L. Sentis are with the Department of Aerospace Engineering and Engineering Mechanics, The University of Texas at Austin, TX, 78712, USA. Email: \{bangsh0718,carlos.gonzalez,eissac412,lsentis\}@utexas.edu}
}

\begin{document}

\maketitle
\thispagestyle{empty}
\pagestyle{empty}

\begin{abstract}
Planning safe motions for legged robots requires sophisticated safety verification tools. However, designing such tools for such complex systems is challenging due to the nonlinear and high-dimensional nature of these systems' dynamics. In this letter, we present a probabilistic verification framework for legged systems, which evaluates the safety of planned trajectories by learning an assessment function from trajectories collected from a closed-loop system. Our approach does not require an analytic expression of the closed-loop dynamics, thus enabling safety verification of systems with complex models and controllers. Our framework consists of an offline stage that initializes a safety assessment function by simulating a nominal model and an online stage that adapts the function to address the sim-to-real gap. The performance of the proposed approach for safety verification is demonstrated using a quadruped balancing task and a humanoid reaching task. The results demonstrate that our framework accurately predicts the systems' safety both at the planning phase to generate robust trajectories and at execution phase to detect unexpected external disturbances.
\end{abstract}

\section{INTRODUCTION}
Safe motion planning for legged systems should be of essential consideration to prevent falling or colliding with obstacles. The main challenge in safe motion planning is to design safety verification tools that accurately evaluate whether a system will satisfy safety constraints while it is stabilized along desired trajectories by using a given feedback controller and without being too conservative.

In this letter, we propose a framework that learns a safety assessment function that can provide probabilistic verification for motion planning. Our framework trains this function using trajectory data. We rollout a number of trajectories using a nominal model and embed them with their safety properties into a low-dimensional space in which we define their safety probabilities. During the execution phase, upcoming desired trajectories are mapped to this low-dimensional space, and the safety probability is estimated before execution. Note that since the safety probability is computed based on the nominal model, there is a reality gap. In order to reduce this gap, we perform an online adaptation process as we collect trajectories during execution. 

\textbf{Related Work:} Recent work on robust motion planning has considered safety verification methods that characterizes funnels around planned trajectories. The authors in \cite{lqr_tree} employed a linear feedback controller and estimated regions of attraction of the closed-loop system by searching Lyapunov functions, and \cite{funnel_lib,zac} showed robust motion planning on aerial robots. A similar new approach, based on Hamilton-Jacobi reachability analysis \cite{fastrack} and contraction theory \cite{ccm}, proposed an offline characterization of tracking error bounds around trajectories. However, these techniques are computationally intensive and limited to a small class of systems, which make it difficult to be deployed for legged robots which are generally modeled as high-dimensional and hybrid system with sophisticated feedback controllers.

Model predictive control (MPC) has shown to be a promising tool to perform dynamic constrained trajectory optimization. In particular, tube-based MPC considers a simple ancillary feedback controller to bind output trajectories around a nominal path and verifies safety satisfactions for all realizations of uncertainties \cite{LANGSON2004125,8619572}. The authors in \cite{tube_mpc_2020} applied this technique to bipedal walking assuming a linear pendulum model and a simple controller. However, computing invariant tubes for highly non-linear and hybrid systems with sophisticated feedback controllers is challenging. The work in \cite{deeptube} proposed to learn distributions of output trajectories in a data-driven manner, which can then be used for safety verification, but the data-efficiency and sim-to-real gap issues have not been addressed for robot deployment.

The studies in \cite{8461237,9663404} considered a Bayesian optimization technique which evaluates planned trajectories executed with a closed-loop controller and use them to find planner parameters. The authors in \cite{policy_modulation,ahn_2020_ral} trained policies using closed-loop systems to generate swing foot trajectories for walking motion. These frameworks make it possible to optimize planner parameters and to design trajectories such that the resulting closed-loop behaviors satisfy safety constraints. However, these verification methods evaluate trajectory safety only at the planning phase, making it difficult to detect unsafe states arising during execution, for instance, due to unexpected disturbances.

\begin{figure}[t]
    \centering
    \includegraphics[width=0.75\linewidth]{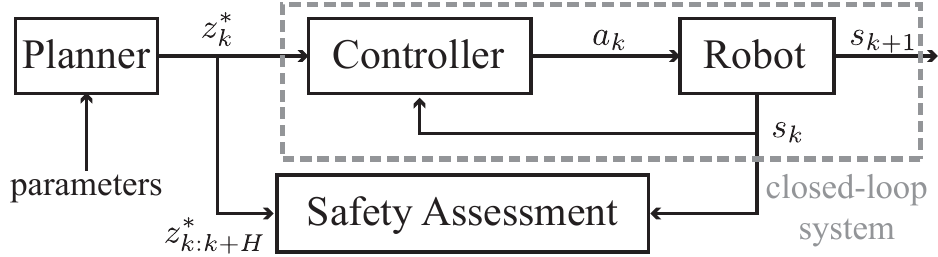}
    \caption{The safety assessment module evaluates the probability of safety of the closed-loop system by taking into account information from the trajectory planner and from the feedback controller.}
    \label{fig:closed_loop_system}
\end{figure}

The idea of embedding system safety information into a low-dimensional space is not new and has been previously presented in \cite{zhou_2021_tnnls}. In this work, the authors proposed a framework that learns a low-dimensional representation of regions of attraction of a closed-loop autonomous system. In our work, we extend this idea and learn a safety assessment function for a closed-loop trajectory tracking system. For closed-loop autonomous systems, the initial states on their own determine the evolution of the systems and therefore, their safety characteristics. On the contrary, closed-loop trajectory tracking systems have external inputs (e.g., desired trajectories), which affect the evolution of the system and, thus, require a special safety treatment. For instance, we have to properly measure which specific pieces of a desired trajectory could result in future failure. To this end, we re-evaluate the computation methods described in \cite{zhou_2021_tnnls} and extend them for safety verification for executing planned trajectories, while preserving algorithmic benefits.

\textbf{Contributions:} Our key contributions are the following: 
\begin{enumerate}[label=(\alph*)]
    \item We propose a framework that learns a safety assessment function that evaluates whether desired trajectories are safe before and during execution. In particular, we investigate a data structure, data generation pipeline, and safety-related properties needed for training.
    \item Our framework incorporates numerous algorithmic advantages, in particular:
    \begin{enumerate}[label=(\roman*)]
        \item It does not require an analytic expression of the closed-loop system to train a safety assessment function, which allows us to reason about safety for complicated systems.
        \item It is data-efficient and is able to address the sim-to-real gap, which is crucial for real system implementation.
        \item Our safety assessment function can provide safety predictions for the trajectories both when generating robust plans and executing to detect unexpected external disturbances.
    \end{enumerate}
    \item We deploy our framework in a quadruped balancing task and a humanoid reaching task and show that our framework can open up a number of interesting possibilities for algorithm development. In the quadruped balancing task, we integrate a back-up recovery step planner that is triggered based on safety predictions, and in the humanoid reaching task, we provide a robot self-assessment capability to estimate the likelihood of safe task completion for human-robot interaction.
\end{enumerate}

\section{PROBLEM STATEMENT}

\begin{figure*}[t]
    \centering
    \includegraphics[width=\linewidth]{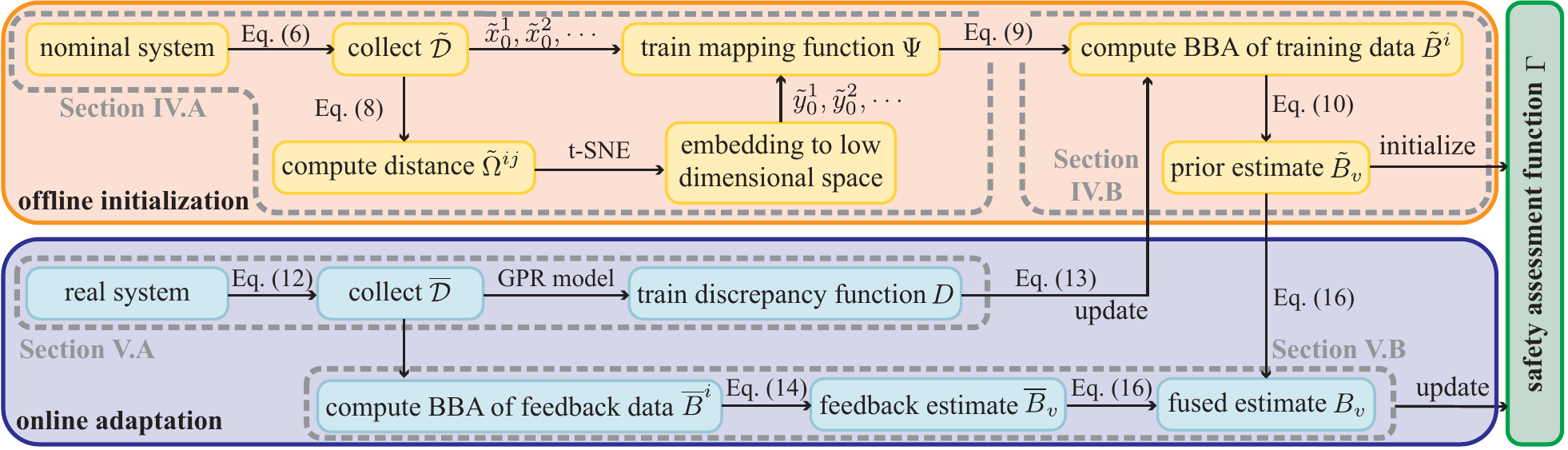}
    \caption{The safety assessment function is initialized through offline process using trajectory data from the nominal system, and then updated through the online adaptation process using trajectory data from the real system to reduce a sim-to-real gap.}
    \label{fig:overview}
\end{figure*}

Consider a discretized system given by
\begin{equation}
\begin{split}
    s_{k+1} &= f(s_k, a_k, w_k), \\
    z_{k} &= g(s_k),
\end{split}
\end{equation}
where $s_k \in \mathbb{R}^{n_s}$, $a_k \in \mathbb{R}^{n_a}$, $w_k \in \mathbb{R}^{n_w}$ are the system state, input, and disturbances. $z_k \in \mathbb{R}^{n_z}$ is the output vector that can be measured from system state (e.g., end-effector positions in task space). We further assume to have a planner that computes a desired trajectory $z^{\ast}_{0:T} \triangleq \begin{bmatrix} {z^{\ast}_{0}}^{\top} & \cdots & {z^{\ast}_{T-1}}^{\top} \end{bmatrix}^{\top}$, where $T$ represents a planning horizon, and $z^{\ast}_{\cdot} \in \mathbb{R}^{n_z}$ denotes a desired output. Given a tracking controller $a_k = K(s_k,z^{\ast}_k)$, the closed-loop system dynamics is denoted as
\begin{equation}
\label{eq:closed_loop_system}
    s_{k+1} = f_{K}(s_k, z^{\ast}_k, w_k) \triangleq f(s_k, K(s_k, z^{\ast}_k), w_k).
\end{equation}
Then, the solution trajectory of the closed-loop system can be recursively computed from the starting state and the upcoming desired trajectory with the expression
\begin{equation}
\label{eq:solution}
    s_{1:T+1} \triangleq \begin{bmatrix} s_1 \\ s_2 \\ \vdots \\ s_T \end{bmatrix} = \begin{bmatrix} f_K(s_0, z_0^{\ast}, w_0) \\ f_K(f_K(s_0, z_0^{\ast}, w_0), z_1^{\ast}, w_1) \\ \vdots \\ f_K(f_K(\cdots), z_{T-1}^{\ast}, w_{T-1}) \end{bmatrix}.
\end{equation}
As illustrated in Fig.~\ref{fig:closed_loop_system}, our goal is to make a receding horizon prediction about the safety of the closed-loop system with the current state measurement and upcoming desired trajectory. To be more specific, at current time index $k$, we want to predict the probability of all future states being safe,
\begin{equation}
\label{eq:saf_output}
    p\left( (s_{k+1} \in S_{\text{safe}}) \cap \cdots \cap (s_{T} \in S_{\text{safe}}) \right),
\end{equation}
using the information of $s_k$ and $z^{\ast}_{k:k+H}$. $S_{\text{safe}}$ is the user-specified safe set that could be defined with a tracking error or conservative capture region to avoid falling. Note that $H$ is the safety assessment horizon during which we look ahead and can be different from the planning horizon $T$. $H$ is a task-dependent parameter and is chosen to contain primarily safety information. For a cyclic walking task, for example, $H$ does not need to be the trajectory duration for multiple steps, but rather just for one stepping cycle. For convenience, we concatenate the state measurement and upcoming desired trajectory and define a safety assessment input:
\begin{equation}
    x_k \triangleq \begin{bmatrix} s_k^{\top} & {z_{k:k+H}^{\ast}}^{\top} \end{bmatrix}^{\top} \in \mathcal{X} \subset \mathbb{R}^{n_s+H n_z}.
\end{equation}
Using this nomenclature, our goal can be summarized to define a safety assessment function $\Gamma:\mathcal{X} \mapsto [0,1]$ that predicts the safety probability \eqref{eq:saf_output} of a closed-loop system.

We consider a scenario where the real dynamical system is not perfectly known, but we assume the nominal system is available and can be simulated over time. Since the dynamics of legged systems are non-linear, high-dimensional, and hybrid and the controller are often formulated based on a numerical optimization problem, we do not have access to the analytic expressions of the closed-loop solution trajectories of either the nominal or real systems. Therefore, we propose to learn the safety assessment function in a data-driven manner. Throughout the paper, we use a tilde, $\tilde{\cdot}$, and an overline, $\overline{\cdot}$, to represent variables related to the nominal system and the real system, respectively.

\section{Framework Overview}
Our framework aims to find a low-dimensional embedding of safety assessment inputs where the low-dimensional space can be discretized into a finite number of grid cells. Then, we assign each cell a belief mass using belief function theory \cite{belief_theory} to evaluate the safety probability of the inputs. The assignment of belief masses is denoted as basic belief assignment (BBA) and the BBA for the grid index $v$ is expressed as $B_v \triangleq ({b}_{v,\text{safe}}, {b}_{v,\text{unsafe}}, \mu_v)$. Here, $b_{v,{\text{safe}}}$ is the belief mass of the probability of the closed-loop system being safe when it evolves with safety assessment inputs that are mapped to and belong to the grid index $v$. $b_{v,{\text{unsafe}}}$ is the belief mass of the complementary event and $\mu_v$ is the uncertainty on the safety estimation. Note that it holds $b_{v,\text{safe}} + b_{v,\text{unsafe}} + \mu_v = 1$, and $b_{v,\text{safe}}$, $b_{v,\text{unsafe}}$, and $\mu_v$ are in the interval $[0,1]$. After the BBAs for the grid cells are computed, we define a safety assessment function $\Gamma(x_k) = b_{v,\text{safe}}$, where the safety assessment input $x_k$ is embedded in the grid cell $v$.

To compute BBAs for grid cells, we first simulate a sufficient amount of trajectories using a nominal model. We collect safety assessment inputs from the trajectories and label them whether they yield safe behaviors or not. For each safety assessment input pair, we evaluate a distance metric to measure their similarity in terms of safety. For instance, the distance between a pair is small if they share a similar safety property (e.g., if they are both safe or unsafe) but large otherwise. Using the computed distances, we embed the safety assessment inputs into a low-dimensional space using the the t-Distributed Stochastic Neighbor Embedding (t-SNE) technique \cite{t-sne}. As a result, we obtain two clusters separated in a low-dimensional space: one is the collection of safety assessment inputs that result in safe behaviors and the other one is the collection of safety assessment inputs that yield unsafe behavior. Then, we discretize the low-dimensional space into grid cells and make a prior estimate of BBA for each cell with the expression $\tilde{B}_v \triangleq (\tilde{b}_{v,\text{safe}}, \tilde{b}_{v,\text{unsafe}}, \tilde{\mu}_v)$.

Simulating the nominal system is usually a cheap and efficient way to initialize the low-dimensional representation of the trajectories and the safety assessment function, but is inaccurate. Therefore, an online adaptation process is followed to reduce the gap between the real and the nominal system and update the safety assessment function. As we collect trajectory data from the real system, we compare it with the behavior from the nominal closed-loop system and train a discrepancy function that reveals how reliable the training data from the nominal system was. Using the discrepancy function, we update the prior estimates of BBAs in the grid cells. At the same time, we compute a feedback estimates of the BBA for each cell using the real system's trajectory data, which is defined as $\overline{B}_v \triangleq (\overline{b}_{v,\text{safe}}, \overline{b}_{v,\text{unsafe}}, \overline{\mu}_{v})$. Finally, we combine the prior and the feedback estimates of BBAs and update the safety assessment function. The overall framework including offline initialization and online adaptation is illustrated in Fig.~\ref{fig:overview}.

\section{OFFLINE INITIALIZATION OF SAFETY ASSESSMENT FUNCTION}

\subsection{Data Generation and Low-dimensional Embedding}
\begin{figure}[t]
    \centering
    \includegraphics[width=\linewidth]{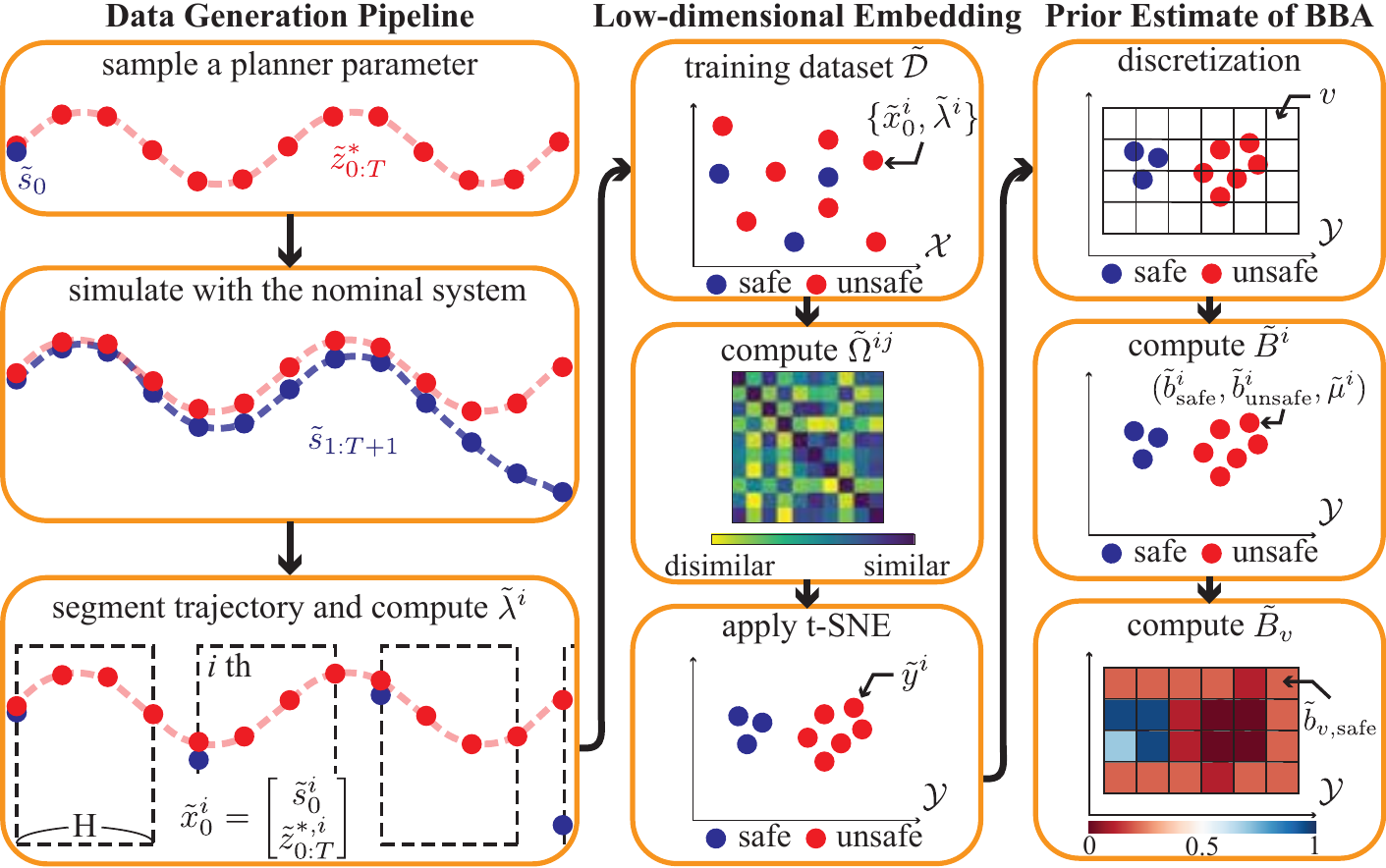}
    \caption{Detailed view of the offline initialization process.}
    \label{fig:offline_initialization}
\end{figure}

As illustrated in Fig.~\ref{fig:offline_initialization}, a planner designs a desired trajectory ($\tilde{z}^{\ast}_{0:T}$) using a randomly sampled planner parameter. Employing a feedback tracking controller, we simulate a nominal closed-loop system and rollout a trajectory ($\tilde{s}_{1:T+1}$). We determine the trajectory to be safe if all of its states are contained in the safe region. We terminate the episode when the system reaches unsafe regions and determine the trajectory to be unsafe. We split the simulated trajectories into segments spanning a duration of $H$, the safety assessment horizon, and create a training data set with each segment's initial state, desired trajectory, and unsafety score. The collection of training data is denoted as $\tilde{\mathcal{D}} = \{ \tilde{D}^i\}^{n_t}_{i=1}$, where
\begin{equation}
        \tilde{D}^i \triangleq \left\{\tilde{x}^i_0, \tilde{\lambda}^i \right\} = \left\{ \begin{bmatrix} \tilde{s}^i_0{}^{\top} & \tilde{z}^{\ast, i}_{0:H}{}^{\top} \end{bmatrix}^{\top}, \tilde{\lambda}^i \right\},
\end{equation}
and $n_t$ is the number of training data, corresponding to the number of trajectory segments. $\tilde{s}^i_0$ and $\tilde{z}^{\ast, i}_{0:H}$ represent the starting state and the desired trajectory of the $i$th trajectory segment \--- note that we zero the beginning time index for each segment \--- forming the $i$th safety assessment input. $\tilde{\lambda}^{i}$ is the unsafety score and is computed by the following rule:
\begin{equation}
\label{eq:unsafety_score}
    \tilde{\lambda}^{i} = \begin{cases} 0, & \text{if the $i$th trajectory is safe} \\ \gamma^{R(i)}, & \text{otherwise} \end{cases},
\end{equation}
where $\gamma \in [0,1]$ is a discount factor and $R:\mathbb{N} \mapsto \mathbb{N}$ is a function that takes a segment index and returns the remaining time steps from the beginning of the segment to the termination of the episode where the segment belongs to. Note that the tilde conveys that the unsafety score is evaluated using the simulated trajectory from the nominal closed-loop system. The unsafety score represents how much the segment contributes to the system's unsafe behavior. Associating it with the discount factor, the segments that are near the episode termination are scored with higher values.

For each pair of training data, we measure their similarity based on their error and safety properties. First, we measure the dynamic time warping for the error signals between the $i$th and $j$th training data using the formula $\tilde{w}^{ij} = \text{DTW}(\tilde{e}^i_{0:H}, \tilde{e}^j_{0:H})$, where $\tilde{e}^{\cdot}_{0:H} \triangleq \tilde{z}^{\ast,\cdot}_{0:H} - \tilde{z}^{\cdot}_{0:H}$ is the trajectory error and $\text{DTW}(\cdot, \cdot)$ is the dynamic time warping operator. While a dynamic time warping measurement might reflect similarity of the safety property in general, it is still possible that safe and unsafe segments share similar trajectories. To obtain more accurate similarity measures in terms of safety, we propose a distance metric considering the dynamic time warping measurements and unsafety scores at the same time as
\begin{equation}
    \tilde{\Omega}^{ij} = \frac{\tilde{w}^{ij}}{\tilde{w}_{\text{max}}} + \delta_{\tilde{\lambda}}|\tilde{\lambda}^{i} -  \tilde{\lambda}^{j}|,
\end{equation}
where $\tilde{w}_{\text{max}}$ denotes the maximum value among the dynamic time warping measurements and $\delta_{\tilde{\lambda}}$ is a weighting constant multiplying the unsafety score difference. As a result, the trajectory segments which show similar error sequences and are alike in terms of safety are considered to be close.

Using this computed distance, we apply t-SNE on the training data to obtain a realization of the low-dimensional space $\mathcal{Y} \subset \mathbb{R}^{n_y}$. Based upon this embedding, we train a mapping function $\Psi:\mathcal{X} \mapsto \mathcal{Y}$, using a deep neural network by minimizing the cost function $||\tilde{y}^i -\Psi(\tilde{x}^i_0)||_2$, where $\tilde{y}^i \in \mathcal{Y}$ is the low-dimensional embedding of the $i$th training data, $\tilde{D}^{i}$. The neural network is trained to reproduce the low-dimensional embedding constructed by t-SNE.

\subsection{Prior Estimate of BBAs on Grid Cells}
We discretize the low-dimensional space into grid cells and compute a prior estimate of BBA for each cell as illustrated in Fig.~\ref{fig:offline_initialization}. For convenience, we define a locating function $L:\mathcal{X} \mapsto \mathbb{Z}^{n_y}$ which takes a safety assessment input and returns an index of a grid cell in which the input is embedded in the low-dimensional space. First, we define the belief assignment for each embedded training data point, $\tilde{y}^i$, based on its unsafety score by introducing the expression $\tilde{B}^i \triangleq (\tilde{b}^i_{\text{safe}}, \tilde{b}^i_{\text{unsafe}}, \tilde{\mu}^i)$, where
\begin{equation}
\label{eq:bba_training_data}
\begin{split}
    &\tilde{b}^i_{\text{safe}} = (1-\tilde{\mu}_{\text{ini}})(1-\tilde{\lambda}^i), \\
    &\tilde{b}^i_{\text{unsafe}} = (1-\tilde{\mu}_{\text{ini}})\tilde{\lambda}^i, \\
    &\tilde{\mu}^i = \tilde{\mu}_{\text{ini}}.
\end{split}
\end{equation}
Here, $\tilde{b}^i_{\text{safe}}$ is the belief mass of the probability of the closed-loop system's behavior being safe when it starts at the state $\tilde{s}^i_0$ with the upcoming desired trajectory $\tilde{z}^{\ast,i}_{0:H}$ and $\tilde{b}^i_{\text{unsafe}}$ is the belief mass of its complementary event. $\tilde{\mu}^i$ represents the confidence level on the nominal system model and is set to user-specified parameter, $\tilde{\mu}_{\text{ini}}$.

We take the belief assignments on the training data into account and further designate a belief assignment for each grid cell. Let us define, for each index $v$, a set of BBAs $\tilde{\mathcal{B}}_{v} \triangleq \{ \tilde{B}^i \vert L(\tilde{x}^i_0)=v \}$, which contains the BBAs for grid cell $v$. Then, the prior estimate of the BBA for the grid cell $v$ can be computed as
\begin{align}
\label{eq:compute_prior_estmate_bba}
   \tilde{B}_v = ( \tilde{b}_{v, \text{safe}}, \tilde{b}_{v, \text{unsafe}}, \tilde{\mu}_{v}) = \left\{ \begin{array}{@{}ll@{}} F(\tilde{\mathcal{B}}_v), & \text{if }\tilde{k}_v \geq \tilde{k}_{\text{min}}   \\ B_{\emptyset}, & \text{otherwise} \end{array} \right.
\end{align}
where $\tilde{k}_v$ is the number of BBAs in $\tilde{\mathcal{B}}_v$, $\tilde{k}_{\text{min}}$ is the minimum number of data for the estimate. When there is not sufficient training data in the grid cell $v$ (i.e., $\tilde{k}_v \leq \tilde{k}_{\text{min}}$), we estimate $\tilde{B}_v$ by an empty BBA $B_{\emptyset} \triangleq (0,0,1)$, which indicates that no safety estimate can be made. $F(\cdot)$ is a fusion operator among the set $\tilde{\mathcal{B}}_v$, which is borrowed from \cite{zhou_2021_tnnls} as
\begin{equation}
\label{eq:fuse_op}
    \begin{split}
        &\tilde{b}_{v, \text{safe}} = \frac{\sum_{\tilde{B}^i \in \tilde{\mathcal{B}}_v} \tilde{b}^i_{\text{safe}} (1-\tilde{\mu}^i) \prod_{\substack{{\tilde{B}^j \in \tilde{\mathcal{B}}_v} \\  {i \neq j}}} \tilde{\mu}^j}{ \left( \sum_{\tilde{B}^i \in \tilde{\mathcal{B}}_v} \prod_{\substack{{\tilde{B}^j \in \tilde{\mathcal{B}}_v}\\ {i \neq j}}} \tilde{\mu}^j \right) - \tilde{k}_v \prod_{\tilde{B}^i\in \tilde{\mathcal{B}}_v} \tilde{\mu}^i},\\
        &\tilde{b}_{v, \text{unsafe}} = \frac{\sum_{\tilde{B}^i \in \tilde{\mathcal{B}}_v} \tilde{b}^i_{\text{unsafe}} (1-\tilde{\mu}^i) \prod_{\substack{{\tilde{B}^j \in \tilde{\mathcal{B}}_v} \\  {i \neq j}}} \tilde{\mu}^j}{ \left( \sum_{\tilde{B}^i \in \tilde{\mathcal{B}}_v} \prod_{\substack{{\tilde{B}^j \in \tilde{\mathcal{B}}_v}\\ {i \neq j}}} \tilde{\mu}^j \right) - \tilde{k}_v \prod_{\tilde{B}^i\in \tilde{\mathcal{B}}_v} \tilde{\mu}^i},\\
        &\tilde{\mu}_{v} = \frac{ \left(\tilde{k}_v - \sum_{\tilde{B}^i \in \tilde{\mathcal{B}}_v} \tilde{\mu}^i \right) \prod_{\tilde{B}^i \in \tilde{\mathcal{B}}_v} \tilde{\mu}^i}{ \left( \sum_{\tilde{B}^i \in \tilde{\mathcal{B}}_v} \prod_{\substack{{\tilde{B}^j \in \tilde{\mathcal{B}}_v}\\ {i \neq j}}} \tilde{\mu}^j \right) - \tilde{k}_v \prod_{\tilde{B}^i\in \tilde{\mathcal{B}}_v} \tilde{\mu}^i}.
    \end{split}
\end{equation}
Finally, the safety assessment function $\Gamma$ is initialized with the prior estimate of the BBAs for grid cells.

\section{ONLINE ADAPTATION OF SAFETY ASSESSMENT FUNCTION}

\subsection{Discrepancy Function}
\begin{figure}[t]
    \centering
    \includegraphics[width=\linewidth]{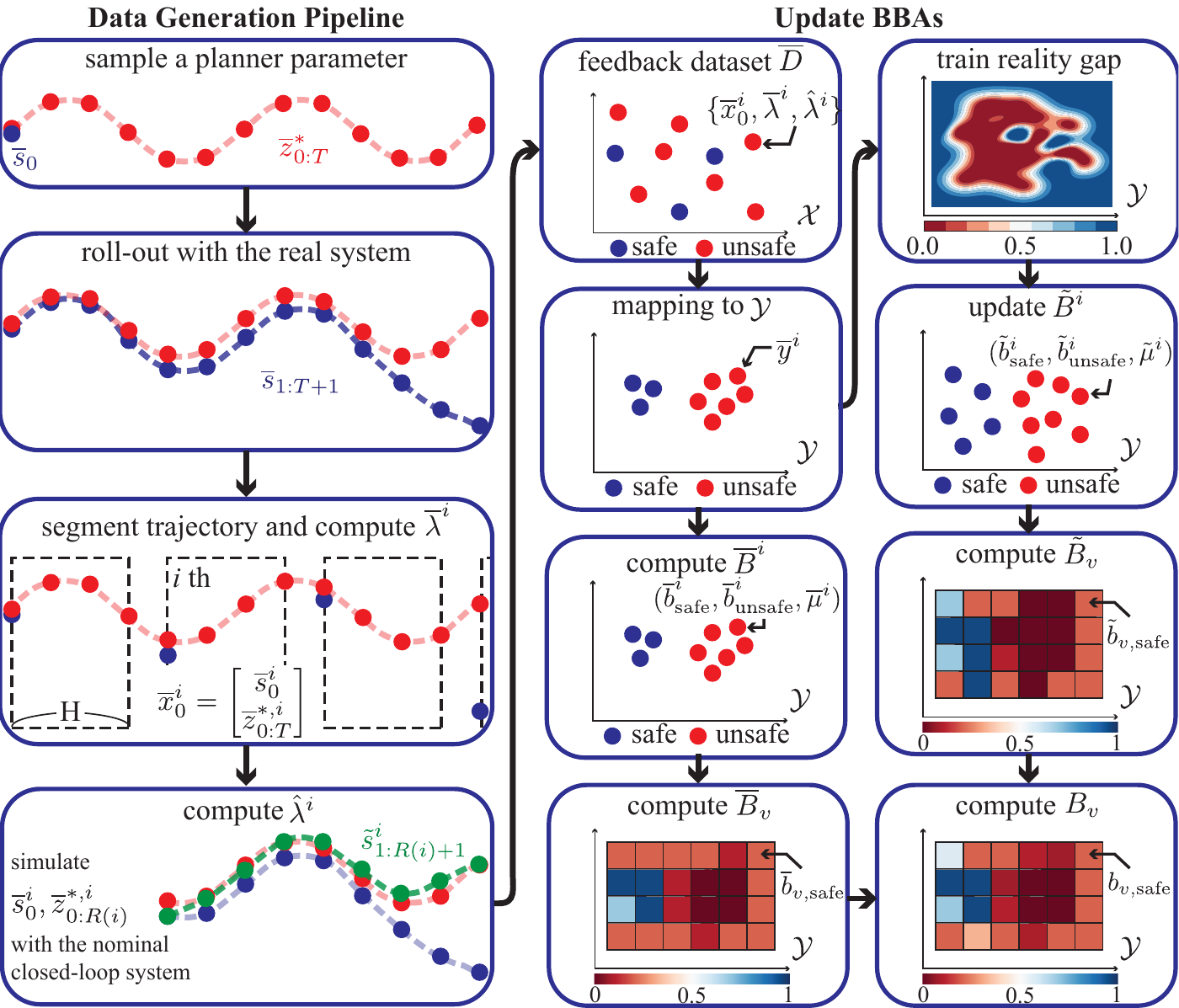}
    \caption{Detailed view of the online adaptation process.}
    \label{fig:online_adaptation}
\end{figure}

Although the prior estimate of the BBA provides a rough safety prediction, we update the safety assessment function online as we collect trajectory data from the real system as depicted in Fig.~\ref{fig:online_adaptation}. When we rollout a trajectory using the real system, we simulate a trajectory using the nominal closed-loop system with the same initial state and the same desired trajectory. With the trajectories from the real and nominal systems, we construct a collection of feedback data $\overline{\mathcal{D}}=\{ \overline{D}^i\}_{i=1}^{n_f}$ with $n_f$ sets, where
\begin{equation}
\label{eq:feedback_dataset}
    \overline{D}^i \triangleq \left\{ \overline{x}^i_0, \overline{\lambda}^i, \hat{\lambda}^i \right\} = \left\{ \begin{bmatrix} \overline{s}^i_0{}^{\top} & \overline{z}^{\ast,i}_{0:H}{}^{\top} \end{bmatrix}^{\top}, \overline{\lambda}^i, \hat{\lambda}^i\right\}.
\end{equation}
Similar to the training data, $\overline{s}^i_0$ and $\overline{z}^{\ast,i}_{0:H}$ represent the starting state and the desired trajectory of the $i$th trajectory segment with the re-ordered time index. $\overline{\lambda}^i$ and $\hat{\lambda}^i$ are the unsafety scores of the $i$th segment of the trajectories of the real and the nominal system, respectively, computed by Eq.~\eqref{eq:unsafety_score}. If there is a discrepancy in terms of safety between the nominal and the real system due to the reality gap, $\overline{\lambda}^i$ can be different from $\hat{\lambda}^i$.

Now, we define a discrepancy function $D: \mathcal{Y} \mapsto [0, 1]$ that quantifies the level of reality gap. We approximate this function with a Gaussian process regression (GPR) model, which is trained with the input set $\{\Psi(\overline{x}^i_0) \}_{i=1}^{n_f}$ and the output set $\{ \vert \overline{\lambda}^i - \hat{\lambda}^i \vert \}_{i=1}^{n_f}$.

With the trained GPR model, we predict the reliability of the training data $\tilde{\mathcal{D}}$ and update the prior estimate of BBA $\tilde{B}_v$. Let us denote the predicted mean and standard deviation of $\tilde{y}^i$ by $m(\tilde{y}^i)$ and $\sigma(\tilde{y}^i)$. Based on the level of reality gap predicted by the trained GPR model, we update the belief assignment on the training data $\tilde{B}^i$ with the new uncertainty
\begin{equation}
        \tilde{\mu}^i = \begin{cases} \tilde{\mu}_{\text{min}} + m(\tilde{y}^i) (1-\tilde{\mu}_{\text{min}}), & \text{if } \sigma(\tilde{y}^i) \leq \sigma_{\text{thre}}\\ \tilde{\mu}_{\text{ini}}, & \text{otherwise}\end{cases},
\end{equation}
where $\tilde{\mu}_{\text{min}}$ is a user-specified parameter set to be smaller than $\tilde{\mu}_{\text{ini}}$. As more feedback data is collected and the standard deviation on the prediction goes below a certain threshold (i.e., $\sigma(\tilde{y}^i) \leq \sigma_{\text{thre}}$), we update the uncertainty of the belief assignment $\tilde{\mu}^i$ using the mean prediction $m(\tilde{y}^i)$. With the new $\tilde{\mu}^i$, we update the belief mass, $\tilde{b}^i_{\text{safe}}$ and $\tilde{b}^i_{\text{unsafe}}$, by following Eq.~\eqref{eq:bba_training_data}. Finally, we improve the prior estimate of BBAs for grid cells with Eq.~\eqref{eq:compute_prior_estmate_bba} to take the reality gap into account.

\subsection{Feedback Estimate of BBAs on Grid Cells}
We update the feedback estimate of BBAs on grid cells using $\overline{\mathcal{D}}$. We, again, first compute the belief assignment for each embedded feedback data with the expression $\overline{B}^i \triangleq (\overline{b}^i_{\text{safe}}, \overline{b}^i_{\text{unsafe}}, \overline{\mu}^i)$, where $\overline{b}^i_{\text{safe}} = 1-\overline{\lambda}^i$, $\overline{b}^i_{\text{unsafe}} =\overline{\lambda}^i$, and $\overline{\mu}^i = 0$. Note that $\overline{\mu}^i$ is set to have zero uncertainty since it comes from the real system. With this, we compute the feedback estimate of BBA for the grid index $v$ as 
\begin{equation}
    \overline{B}_{v} = \begin{cases} G(\overline{\mathcal{B}}_v), & \text{if } \overline{k}_v \neq 0 \\
    B_{\emptyset}, & \text{otherwise} \end{cases},
\end{equation}
where $\overline{\mathcal{B}}_v \triangleq \{ \overline{B}^i \vert L(\overline{x}^i_0) = v \}$ contains the BBAs in grid $v$, and $\overline{k}_v$ is the number of BBAs in the set $\overline{\mathcal{B}}_v$. If no feedback data is collected yet for the index $v$ (i.e., $\overline{k}_v=0$), we set the estimate to an empty BBA. $G(\cdot)$ is another fusion operator among the set $\overline{\mathcal{B}}_v$ and is defined as
\begin{equation}
    \begin{split}
        &\overline{b}_{v,\text{safe}} = (1-\overline{\mu}_{v})\sum\nolimits_{\overline{B}^i \in \overline{\mathcal{B}}_v}\frac{ \overline{b}^i_{\text{safe}}}{\overline{k}_v} \\
        &\overline{b}_{v,\text{unsafe}} = (1-\overline{\mu}_{v})\sum\nolimits_{\overline{B}^i \in \overline{\mathcal{B}}_v}\frac{ \overline{b}^i_{\text{unsafe}}}{\overline{k}_v}\\
        &\overline{\mu}_{v} = \beta \exp \left(-\alpha(n_f - 1) \right)
    \end{split}
\end{equation}
Here, parameters $\beta$ and $\alpha$ are the initial value and the decay rate of the uncertainty $\overline{\mu}_v$, respectively, and the uncertainty converges to zero as the number of data goes to infinity (i.e., $\lim_{n_f \to \infty} \overline{\mu}_v \to 0$). $\overline{b}_{v,\text{safe}}$ and $\overline{b}_{v,\text{unsafe}}$ are computed with the average operator.

Finally, we combine $\tilde{B}_v$ and $\overline{B}_v$ and compute the BBA for each index vector $v$ as
\begin{equation}
    B_v = \begin{cases}F(\left\{ \tilde{B}_v, \overline{B}_v\right\}), & \text{if } \overline{B}_v \neq B_{\emptyset} \\ \tilde{B}_v, & \text{otherwise}. \end{cases}
\end{equation}
If the feedback estimate for the grid index $v$ is available, we fuse the prior and feedback estimates of BBAs through the fusion operator in Eq.~\eqref{eq:fuse_op}, otherwise, we just use the prior estimate. It has been shown that the $B_v$ approaches $\overline{B}_v$ as the number of feedback data, $n_f$, approaches infinity \cite{zhou_2021_tnnls}. This means that the prior estimate has an effect when there is no sufficient data from the real system, but has less of an effect in making safety estimates. We finally update the safety assessment function as $\Gamma(x_k) = b_{v,\text{safe}}|_{v=L(x_k)}$. For computational efficiency, the online adaptation process is performed once every $k_u$ sets of feedback data are obtained, where the value of $k_u$ is a task dependant parameter.

\section{EXPERIMENTAL RESULTS}
\begin{figure*}[t]
    \centering
    \includegraphics[width=\linewidth]{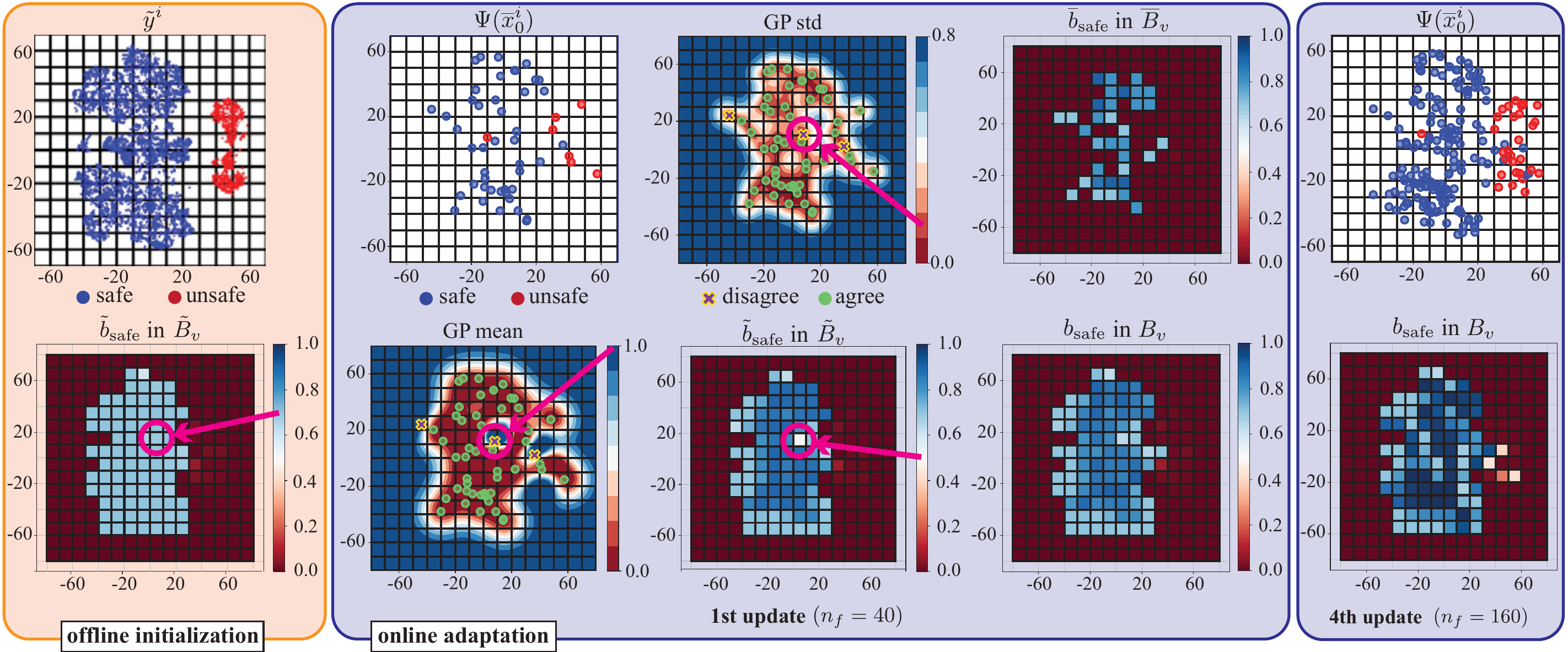}
    \caption{Offline initialization (left) and online adaptation (right) of safety assessment function during Laikago's balancing task. The online adaptation process occurs once every $40$ feedback data sets are collected. For the online adaptation phase, only the first and the fourth iterations are illustrated.}
    \label{fig:result2}
\end{figure*}
\begin{figure*}[t]
    \centering
    \includegraphics[width=\linewidth]{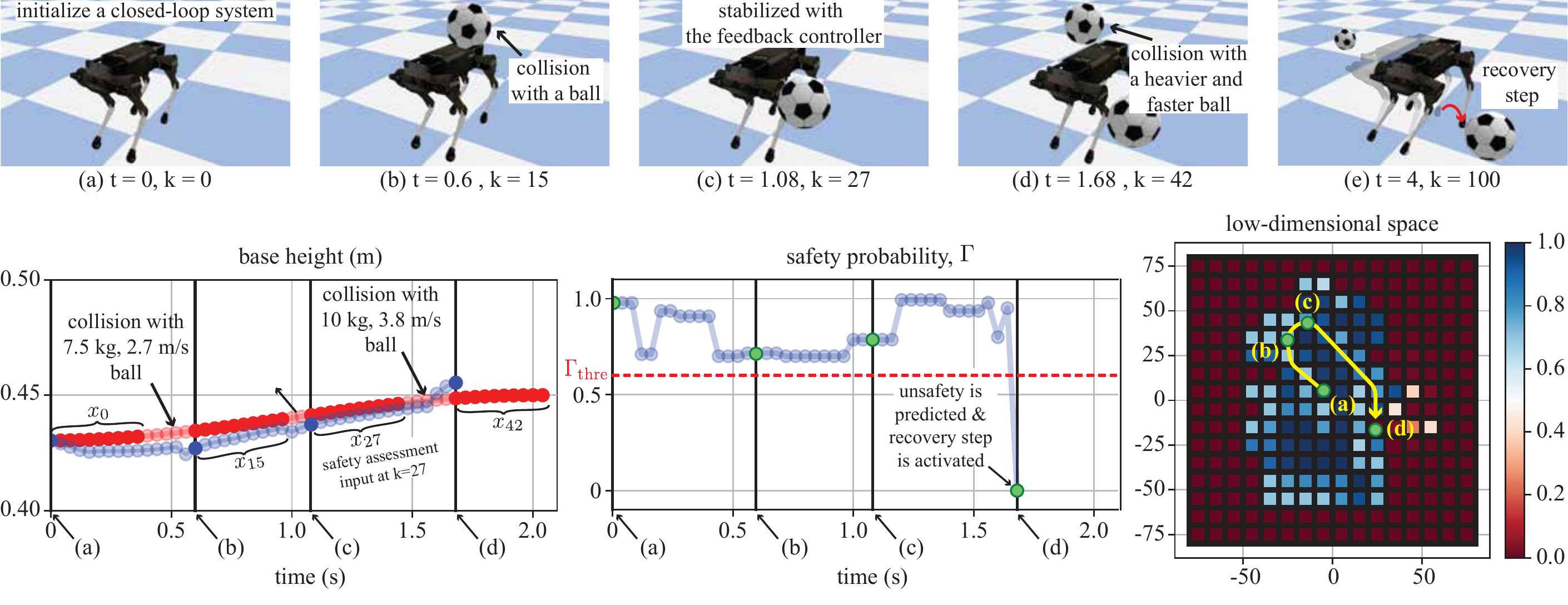}
    \caption{(top) Snapshots of Laikago balancing (a)-(d) and taking a recovery step (e). (bottom) Receding horizon safety prediction over time and throughout the low-dimensional space. The robot is initialized at $k=0$ and is perturbed by the balls twice (at $k=15$ and $k=42$).}
    \label{fig:result1}
\end{figure*}

In this study, we consider two different scenarios: a quadruped balancing task and a humanoid reaching task. We then address the following questions: Does the offline initialization phase find a proper low-dimensional representation of trajectory data and compute $\tilde{B}_v$? Does the online adaptation phase incorporate feedback data and properly address the sim-to-real gap? Can the safety assessment function make a receding horizon prediction so that it can evaluate trajectories' safety both at planning phase and at the execution phase? How is our safety assessment function compared to other baseline verification tools and how much are the predictions accurate? How can our framework be incorporated to a back-up planner or controller to prevent unsafe behaviors?

\begin{figure*}[t]
    \centering
    \includegraphics[width=\linewidth]{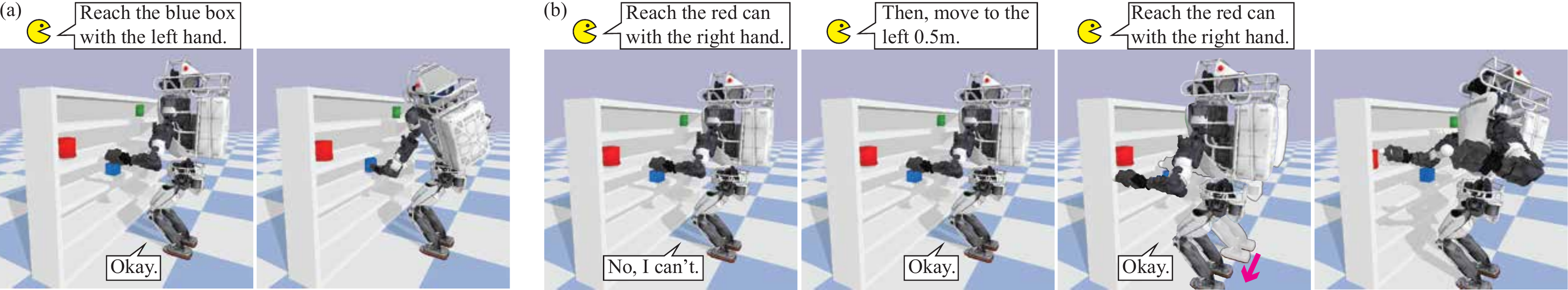}
    \caption{Snapshots of Atlas reaching (a) the blue box and (b) the red can. In these human-robot interaction scenarios, the human tells the robot which object to reach.}
    \label{fig:result3}
\end{figure*}

\begin{figure}[t]
    \centering
    \includegraphics[width=\linewidth]{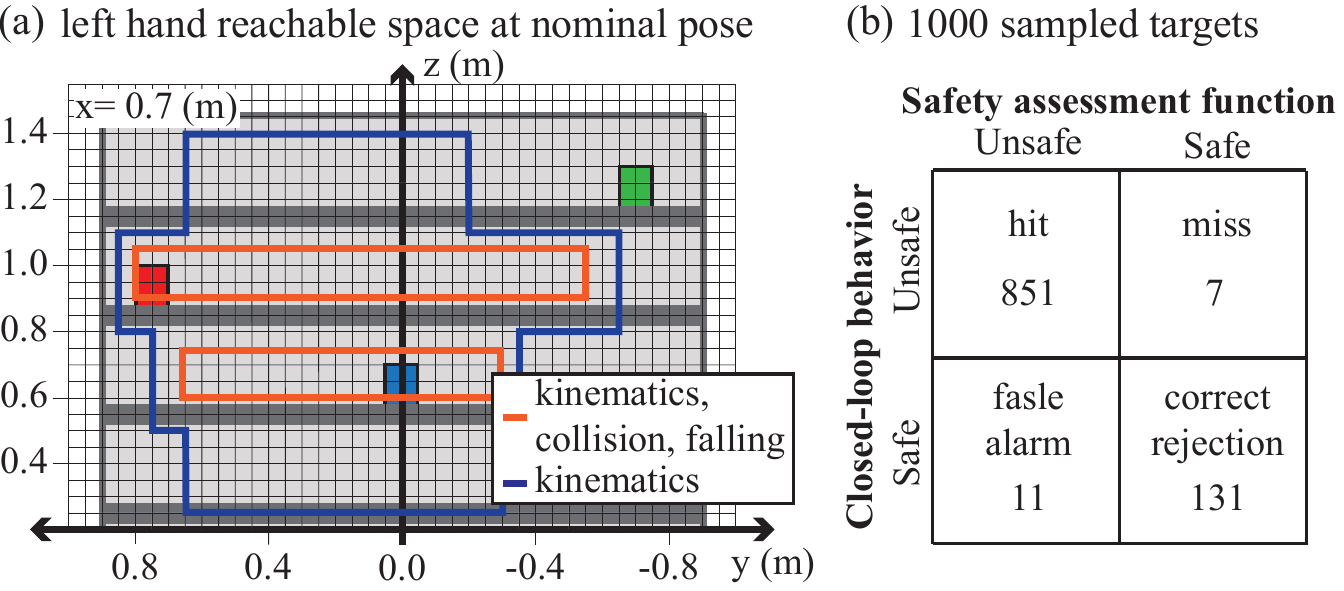}
    \caption{(a) Reachable regions of Atlas' left hand computed by our safety assessment function (orange boundary) and by a simple inverse kinematics-based reachability method (blue boundary) from nominal pose shown in Fig.~\ref{fig:result3}. (b) Safety assessment function predictions based on $1000$ randomly sampled targets and the resulting closed-loop behaviors.}
    \label{fig:result4}
\end{figure}

\subsection{Laikago Balancing}

We consider a balancing task using the Laikago quadruped from UnitreeRobotics. The robot's state $s_k$ consists of its floating base and joints configurations, and the output vector $z_k$ is the base position. At every episode, the robot is initialized with randomly sampled state and our planner generates an interpolated trajectory between the initial and desired base position. Then, our feedback controller computes joint position commands by solving inverse kinematics to follow the trajectory. For this task, we define the safe set $\mathcal{S}_{\text{safe}}$ to be the supporting polygon and a specified height range. Thus, we check that the projection of the base onto the ground remains inside this safe region and that the base height remains within its corresponding bounds. We consider random disturbances while balancing and aim to make a receding horizon safety prediction on the motions using the safety assessment function. If a strong disturbance causing the closed-loop system to become unsafe is properly detected by the safety prediction module, we initiate a recovery step plan \cite{raibert_heuristic} to avoid falling. Table~\ref{table:hyperparameters} summarizes parameters used in the safety assessment function training.

We simulate $2048$ episodes with the nominal closed-loop system and segment the data to construct the training data $\tilde{\mathcal{D}}$.\footnote{We intentionally make a reality gap by reducing the link's mass by $20\%$ and removing the joint frictions and observation noises to simulate the nominal system. We also add a random offset to the initial state to simulate the disturbances.} We measure the distance between the training data and use it to embed the data into a two dimensional space (i.e., $n_y=2$) that is discretized into a $14$ by $14$ square grid with a cell length of $10$. The low-dimensional embedding of the training data $\tilde{y}^i$ and the prior estimate of BBAs for grid cells are illustrated in Fig.~\ref{fig:result2}.

The online adaptation process is performed once every $40$ feedback data are collected from the real system (i.e., $k_u=40$). We train the discrepancy function with the GPR model and update $\tilde{B}_v$ for each grid cell. For instance, the grid cell highlighted with the pink circle in Fig.~\ref{fig:result2} was originally assigned $70\%$ of safety probability in the offline initialization phase but is updated to $50\%$ after the first update iteration due to the feedback data that shows a large sim-to-real gap. This makes the discrepancy prediction around the pink circle regions to be high, which results in an increase in the uncertainty $\tilde{\mu}_v$ and a decrease in the safety probability $\tilde{b}_{\text{safe}}$. At the same time, we update $\overline{B}_v$ and fuse it with $\tilde{B}_v$ to adapt the safety assessment function.

After the safety assessment module converges, we show that our framework can make a receding horizon safety prediction on the balancing trajectories and trigger the recovery step when it is needed to avoid falling. Fig.~\ref{fig:result1} shows snapshots of Laikago balancing and taking a recovery step. The robot is perturbed with balls in simulation: one which generates a small disturbance (Fig.~\ref{fig:result1}(b)) and another one which generates a large disturbance (Fig.~\ref{fig:result1}(d)). The robot stabilizes and tracks the desired trajectory until the safety assessment function predicts future unsafety. When it predicts a safety probability below the threshold $\Gamma_{\text{thre}}$, set to $0.6$, it triggers the recovery step planner to avoid falling.

\subsection{Atlas Reaching}

We consider an object reaching task using the Boston Dynamic's humanoid Atlas. The robot's state $s_k$ consists of its floating base and joints configurations, and the output vector $z_k$ consists of the reaching hand position. At every episode, the robot is initialized with randomly sampled state and the planner generates an interpolated trajectory between the initial and the target hand position. Our feedback controller computes joint torque commands by using an optimization-based whole-body controller \cite{2021_ahn_froniter}. We define the safety set such that $s_k \in \mathcal{S}_{\text{safe}}$ if the projected base position is inside the supporting polygon, the end-effectors do not collide with the obstacles, and the joint positions remain within their limits. We train the safety assessment function for the hand reaching trajectories and use it to predict whether the robot can reach the commanded target safely.\footnote{When we rollout trajectories using the nominal system, we do not sample an offset and do not add it to the initial state since we do not consider disturbances here.} This training is done only for one arm since the same mapping function can be used for both left and right arms. The parameters used in the training are identical to the ones used in Laikago balancing task except for the prediction horizon, which is $30$.

\begin{table}[t]
\caption{Parameters}
\label{table:hyperparameters}
\centering
\begin{tabular}{ccccccccc}
\hline
$\gamma$ & $H$ & $\delta_{\tilde{\lambda}}$ & $\tilde{\mu}_{\text{ini}}$ & $\tilde{k}_{\text{min}}$ & $\sigma_{\text{thre}}$ & $\tilde{\mu}_{\text{min}}$ & $\alpha$ & $\beta$ \\ \hline\hline
0.99     & 10  & 0.01                       & 0.3                        &  5                        & 0.3                    & 0.1                & 0.4                     & 0.3    \\ \hline
\end{tabular}
\end{table}

When a human commands a humanoid what to do as an end-user, it is not trivial to evaluate whether the command is safe to execute or not. We demonstrate that our safety assessment function enables a robot to estimate the likelihood it will accomplish the given task safely. Fig.~\ref{fig:result3}(a) illustrates a scenario where Atlas is told to reach the blue box on the bookshelf. After ensuring this task can be accomplished safely, the robot executes the command. Fig.~\ref{fig:result3}(b) illustrates the scenario where the robot is initially told to reach the red can. Based on the safety prediction, the robot rejects the task so that the human instructor can provide a different description to accomplish the task.

In Fig.~\ref{fig:result4}(a), we compare the reachable regions on the bookshelf computed by our safety assessment function against those obtained by a simple inverse kinematics based reachability method. Our safety assessment function considers joint limits violation, collision, and falling down while manipulating to be unsafe, and it results in more conservative reachable regions than those considering only kinematic constraints. Fig.~\ref{fig:result4}(b) summarizes the evaluation on the prediction accuracy of our safety assessment function. Among $1000$ episodes with randomly sampled target positions, the safety assessment function predicts $95.2\%$ of safe targets to be safe and $98.7\%$ of unsafe targets to be unsafe.

\section{CONCLUSIONS}

In this letter, we propose a probabilistic safety verification tool for legged systems when desired motions are given. We leverage a low-dimensional embedding of the current state measurement and upcoming desired trajectories based on the proposed distance metric for safety prediction. For data-efficiency, we initialize our safety assessment function by simulating trajectories with a nominal system and perform online adaptation using trajectories from the real system to account for the reality gap. We have demonstrated our framework's efficiency and accuracy with a quadruped balancing task and a humanoid reaching task. 

As future work, we would like to integrate our safety verification tool in hierarchical reinforcement learning frameworks such as \cite{option} and train a high-level motion policy with a safety consideration. We would also like to deploy our safety verification tool in a human-robot interaction scenario such as \cite{hri} and provide self-assessment capabilities to our new Draco humanoid, a successor of the Draco biped \cite{2019_ahn}.

\addtolength{\textheight}{-12cm}   




\section*{ACKNOWLEDGMENT}
The authors would like to thank the members of the Human Centered Robotics Laboratory at The University of Texas at Austin for their great help and support.

\bibliographystyle{IEEEtran}
\bibliography{references}

\end{document}